\documentclass[10pt,twocolumn,letterpaper]{article}

\usepackage{iccv}
\usepackage{times}
\usepackage{epsfig}
\usepackage{graphicx}
\usepackage{amsmath}
\usepackage{amssymb}

\usepackage{multirow}
\usepackage{booktabs}

\def\methodName{3DSNet}
\def\methodNameMultimodal{3DSNet-M}
\def\adanorm{AdaNorm}
\def\adabn{AdaBN}
\def\adain{AdaIN}
\def\metricName{STS}
\def\lpips{LPIPS}
\def\threedlpips{3D-LPIPS}

\def\meshflow{MeshFlow}
\def\atlasnet{AtlasNet}
\def\pointnet{PointNet}
\def\psnet{PSNet}

\def\shapenet{ShapeNet}
\def\smal{SMAL}

\def\chairs{Chairs}
\def\planes{Planes}

\def\chairsPair{\textit{armchair} $\leftrightarrow$ \textit{straight chair}}
\def\planesPair{\textit{jet} $\leftrightarrow$ \textit{fighter aircraft}}

\def\hippos{\textit{hippos}}
\def\horses{\textit{horses}}
\def\smalPair{\hippos{} $\leftrightarrow$ \horses{}}

\usepackage{pifont}
\newcommand{\cmark}{\ding{51}}%
\newcommand{\xmark}{\ding{55}}%

\def\domainOne{\mathcal{X}_1}
\def\domainTwo{\mathcal{X}_2}
\def\shapeOne{x_1}
\def\shapeTwo{x_2}
\def\shapeOneTwo{x_{1 \rightarrow{} 2}}

\def\contentSpace{\mathcal{C}}

\def\styleSpace{\mathcal{S}}

\def\contentEncoder{E^c}
\def\styleEncoder{E^s}
\def\globalDecoder{H}
\def\contentDecoder{G}
\def\mappingFunction{M}
\def\discriminator{D}

\def\pOneTwo{P_{\mathcal{X}_1,\mathcal{X}_2}}
\def\pOne{P_{\mathcal{X}_1}}
\def\pTwo{P_{\mathcal{X}_2}}
\def\pOneCondTwo{P_{\mathcal{X}_1|\mathcal{X}_2}}
\def\pTwoCondOne{P_{\mathcal{X}_2|\mathcal{X}_1}}
\def\pLearnedOneCondTwo{P_{\mathcal{X}_{2 \rightarrow{} 1}|\mathcal{X}_2}}
\def\pLearnedTwoCondOne{P_{\mathcal{X}_{1 \rightarrow{} 2}|\mathcal{X}_1}}

\def\identity{\mathrm{\mathbf{I}}}

\usepackage[pagebackref=true,breaklinks=true,letterpaper=true,colorlinks,bookmarks=false]{hyperref}

\usepackage{cleveref}  

\iccvfinalcopy 

\def\iccvPaperID{5717} 

\ificcvfinal\fi

\begin{document}

\title{\methodName{}: Unsupervised Shape-to-Shape 3D Style Transfer}

\author{Mattia Segu$^{1}$,  Margarita Grinvald$^{1}$,  Roland Siegwart$^{1}$,  Federico Tombari$^{2,3}$\\	
		\tt\small \{segum, mgrinvald, rsiegwart\}@ethz.ch, tombari@google.com \\
		\hspace{0.7cm}
		$^{1}$ ETH Zurich \hspace{1.5cm} 
		$^{2}$ TU Munich \hspace{1.5cm} 
		$^{3}$ Google \hspace{1.2cm}
		\\
	}

\maketitle
\ificcvfinal\thispagestyle{empty}\fi

%
\begin{abstract}
Transferring the style from one image onto another is a popular and widely studied task in computer vision. 
%
%
Yet, style transfer in the 3D setting remains a largely unexplored problem.
To our knowledge, we propose the first learning-based approach for style transfer between 3D objects based on disentangled content and style representations.
%
%
%
The proposed method can synthesize new 3D shapes both in the form of point clouds and meshes, combining the content and style of a source and target 3D model to generate a novel shape that resembles in style the target while retaining the source content.
%
%
Furthermore, we extend our technique to implicitly learn the multimodal style distribution of the chosen domains.
By sampling style codes from the learned distributions, we increase the variety of styles that our model can confer to an input shape.
Experimental results validate the effectiveness of the proposed 3D style transfer method on a number of benchmarks.
The implementation of our framework will be released upon acceptance.
\end{abstract}
%
\section{Introduction} \label{sec:introduction}
\begin{figure}[ht!]
\begin{center}
    \includegraphics[scale=0.52,trim={0.0cm 1.0cm 13.0cm 0.0cm},clip]{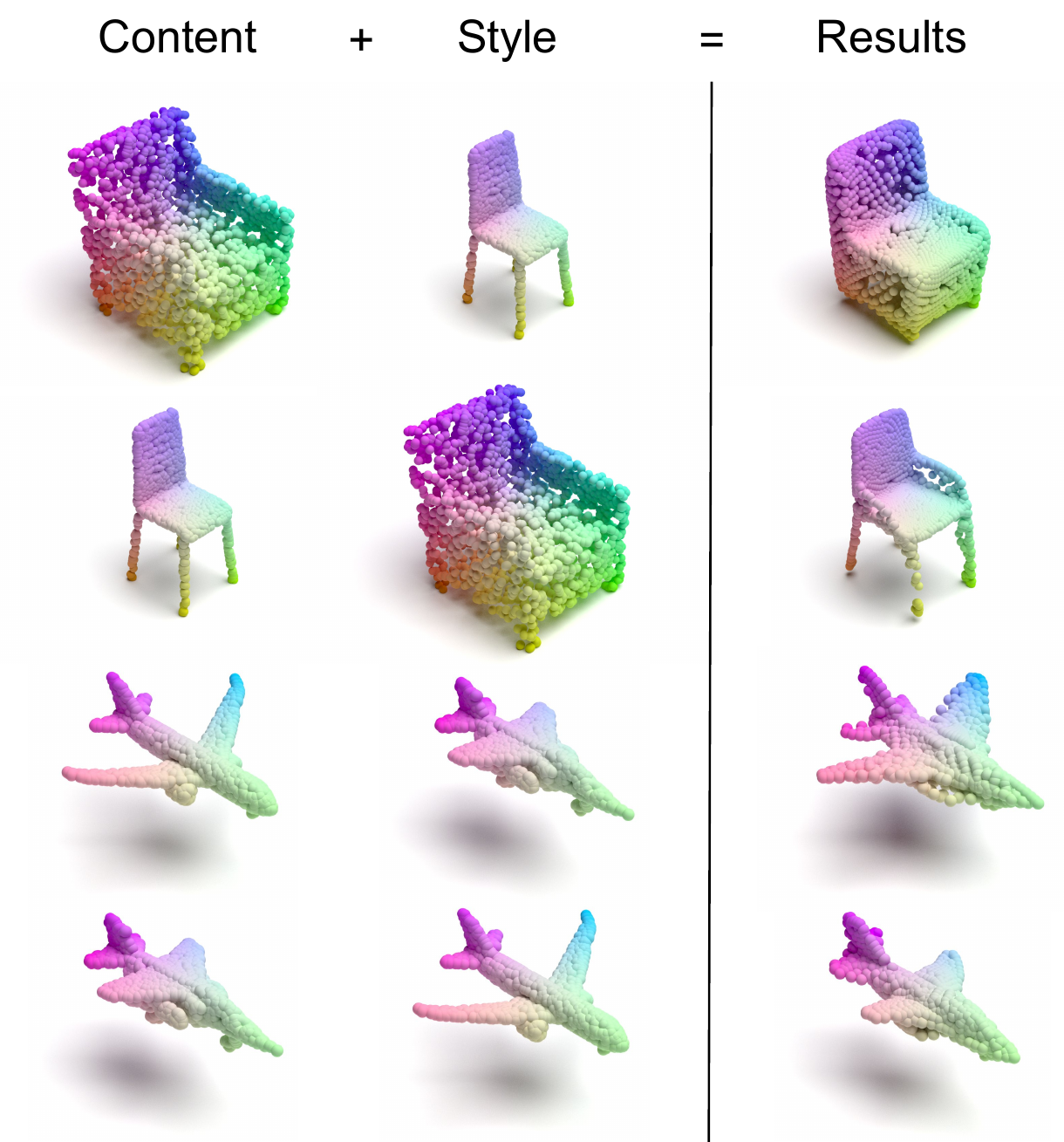} 
\end{center}
\vspace{-2mm}
    \caption{\methodName{} learns to combine the content of one shape with the style of another. We illustrate style transfer results for the pairs \chairsPair{} and \planesPair{}.}
    \label{img:teaser}
\end{figure}


2D style transfer is a widely studied computer vision task, where the style of a reference image is synthesized and applied to another one under the constraint of preserving the semantic content of the latter~\cite{gatys2016image}. 
%
Analogously, transferring the style between 3D objects can be interpreted as transferring the peculiar traits of a target shape to a source shape while preserving the underlying pose and structure of the latter.
In both cases, we refer to the underlying global structure of an image/shape as \textit{content}, and identify as \textit{style} the fine traits that differentiate an element from others with the same content.

While one can think of different interpretations of the concept of style for 3D shapes, we borrow a commonly used definition~\cite{xu2010style,lun2015elements,dosovitskiy2015learning,dosovitskiy2016learning} from the computer graphics community, where style similarity between objects is defined as the presence of similarly shaped, salient, geometric elements~\cite{lun2015elements}.
Since shape parts play a
fundamental role in analyzing a set of shapes~\cite{xu2010style}, one could for example divide the chair category into different style families, \eg straight chairs and armchairs~\cite{xu2010style,dosovitskiy2015learning,dosovitskiy2016learning}.
While the two groups share the same semantic meaning, each is characterized by a set of distinctive traits, such as the anatomy of the backrest or the presence of armrests.
Transferring the style from an armchair to a straight chair thus requires to generate a realistic version of the latter inheriting the armrests from the former.
%

Among the multitude of application domains that 3D style transfer targets, augmented and virtual reality is perhaps the most prominent one.
One can envision users bringing their own furniture items in the virtual world, automatically styled according to a desired theme.
Further, 3D shape translation would enable users to manipulate template avatars~\cite{yifan2020neural} and to create custom ones by simply providing a 3D object as style reference, thus enhancing social presence in virtual gaming and collaboration environments.
In general, interior designers and content creators would be able to generate countless new 3D models by simply combining the content and style of previously existing items.

%
The complexity of the 3D style transfer problem has been tackled by reducing it to the simpler tasks of learning pointwise shape deformations~\cite{groueix2019unsupervised,hanocka2018alignet,wang20193dn,yifan2020neural} or parameterizations of template models~\cite{marin2020farm,cosmo2020limp}.
%
%
Only recently the complete problem has been approached using optimization techniques~\cite{cao2020psnet}.
In contrast, we propose to directly address the 3D style transfer problem by learning style-dependent generative models.
By leveraging an autoencoder architecture, we simultaneously learn reconstruction and style-driven transformations of the input 3D objects.
To realize the transformation, we rely on the properties of normalization layers~\cite{ioffe2015batch, ulyanov2016instance} to capture domain-specific representations~\cite{segu2020batch}.
To this end, we adopt custom versions of the \atlasnet{}~\cite{groueix2018papier} and \meshflow{}~\cite{gupta2020neural} decoders, in which adaptive normalization layers~\cite{ulyanov2016instance,li2018adaptive} replace the standard counterparts to generate style-dependent output distributions.
This choice of decoders allows our framework to generate both point clouds and meshes.
A combination of reconstruction and adversarial losses is used to enhance the output faithfulness and realism.
Moreover, we empower our framework by modifying the training scheme to also learn the multimodal distribution in the style spaces, allowing the learned translation functions to be non-deterministic.
As a consequence, this multimodal version enables to transform a single 3D shape to a larger variety of outputs in the target domain.

To the best of our knowledge, we are first to propose a principled learning-based probabilistic approach to tackle the 3D style transfer problem providing disentangled content and style representations.
%
Our contributions can be summarized as follows: 
%
(i) \methodName{}, a generative learning-based 3D style transfer approach for point clouds and meshes; 
\mbox{(ii) \methodNameMultimodal{}}, the first multimodal shape translation network;
(iii) \threedlpips{} and \metricName{}, two novel metrics designed to measure, respectively, 3D perceptual similarity and 
the effectiveness of style transfer. 
Experimental results on different datasets demonstrate qualitatively and quantitatively how our approach can effectively perform 3D style transfer and generalize across a variety of object categories. 

%
\section{Related Work} \label{sec:related_work}
While style transfer has been widely investigated for images~\cite{gatys2016image,isola2017image,liu2017unsupervised,zhu2017unpaired}, transferring the style across 3D objects~\cite{cao2020psnet,yin2019logan} has not been extensively explored yet. 
We first review the style transfer literature for images (\Cref{ssec:image_style}), and then discuss how current 3D shape processing methods and related 3D shape generation works pave the way towards new 3D style transfer approaches (\Cref{ssec:towards_3d_style}).

\subsection{Image Style Transfer} \label{ssec:image_style}
Style transfer has been extensively studied for images~\cite{gatys2016image,isola2017image,liu2017unsupervised,zhu2017unpaired}.
Preliminary work focused on identifying \textit{explicit} content and style representations.
Gatys~\etal~\cite{gatys2016image} advance the original idea of synthetizing a new image by combining the content representation of one image with the style representation of another.
They propose to extrapolate the content and style representations as, respectively, the intermediate activations and the corresponding Gram matrices of a pre-trained VGG~\cite{simonyan2014very} network. 
The resulting image is generated by iteratively updating a random input until its content and style representations match the reference ones. 

More recent approaches rely instead on adversarial training~\cite{goodfellow2014generative} of generative networks to learn \textit{implicit} representations from two given domains.
In this setting, the main challenge is aligning the two style families.
Some works~\cite{isola2017image,johnson2016perceptual,ledig2017photo,liu2016coupled} propose to learn a deterministic mapping between an image from the source domain and another from the target domain.
Other methods instead are unsupervised~\cite{zhu2017unpaired,liu2017unsupervised,park2020contrastive,lira2020ganhopper}, meaning that no pair of samples from the source and target domain is given.
This is made possible by the cycle-consistency constraint~\cite{zhu2017unpaired,kim2017learning}, enforcing that translating an image to another domain and back should result in the original image.
The image-to-image translation problem has also been tackled in a style-transfer fashion by either combining the content representation of an image with the style of another~\cite{liu2017unsupervised} or randomly sampling the style code from a learned style space~\cite{huang2018multimodal,patashnik2020balagan,lee2018diverse,nizan2020breaking}.
Approaches from the latter category are often referred to as \textit{multimodal}, since they learn the underlying distributions of each style family. This significantly increases the variety of styles that the model can confer to a given content image.

\subsection{Towards 3D Style Transfer} \label{ssec:towards_3d_style}
%
Very few works have dealt with style transfer for 3D objects so far, most of which reduce the 3D shape-to-shape translation problem to a simpler one~\cite{groueix2019unsupervised,hanocka2018alignet,wang20193dn,yifan2020neural,marin2020farm,cosmo2020limp}. 

\noindent\textbf{3D Reconstruction.}
As point clouds and meshes become the most popular representations for several vision and graphics applications, different approaches have been proposed to compress and reconstruct 3D models.
Current shape processing approaches~\cite{groueix2018papier,yang2018foldingnet,yang2019pointflow,gupta2020neural} often adopt an autoencoder architecture, where the encoder projects the input family into a lower-dimensional latent space and the decoder reconstructs the data from the latent embedding.

Being point clouds an irregular data structure, \pointnet{}~\cite{qi2017pointnet} has proven to be an effective encoder architecture. By using symmetric operations (max pooling) shared across all points in a multi-layer perceptron (MLP), it aggregates a global feature vector respecting the permutation invariance of points in the input.
FoldingNet~\cite{yang2018foldingnet} and \atlasnet{}~\cite{groueix2018papier} have both been proposed to reconstruct 3D surfaces by using latent features learned with \pointnet{} to guide folding operations on a set of shape primitives.
%
%
More recently, continuous normalizing flows were adopted in probabilistic frameworks to generate 3D point clouds or meshes~\cite{yang2019pointflow, gupta2020neural}.

\noindent\textbf{3D Generative Models.}
The first model capable of generating point clouds was introduced by Achlioptas~\etal~\cite{achlioptas2018learning}. 
Compared to reconstruction approaches, they adopt an additional adversarial loss to train an autoencoder architecture, leveraging \pointnet{} as encoder and an MLP as decoder.
A follow-up work by Li~\etal~\cite{li2018point} proposes to use two generative models to learn a latent distribution and produce point clouds based on the latent features.
Similarly, StructureNet~\cite{mo2019structurenet} leverages a hierarchical graph network to process meshes.
However, none of these techniques allows to explicitly convert shapes from one domain to another.

\noindent\textbf{Learning 3D Deformations.}
Traditional mesh deformation methods rely on a sparse set of control points~\cite{sumner2004deformation,sorkine2007rigid} or a coarse cage mesh~\cite{thiery2012cager,sacht2015nested,calderon2017bounding,ju2005mean}, whose vertices transformations are interpolated to deform all remaining points.
Recently, learning-based approaches emerged to apply detail-preserving deformations on point clouds or meshes.
Some methods encode an input pair into a latent space and then condition the deformation field of the source shape on the latent code of the target shape~\cite{groueix2019unsupervised,hanocka2018alignet,wang20193dn,gao2018automatic}.
Neural cages~\cite{yifan2020neural} have been proposed to learn detail-preserving shape deformations to resemble the general structure of another reference 3D object.
While these methods excel in preserving details, they only allow pointwise or cage-based deformations limited to the spatial extent, hence lacking the capability of transferring the style of a shape to another.

\noindent\textbf{3D Style Transfer.}
The computer graphics community widely investigated the concept of style for 3D shapes, relating stylistic similarity to the presence of similarly shaped, salient, geometric elements and parts~\cite{xu2010style,lun2015elements}. 
Several approaches have been proposed to analyse the style of a shape, mostly attempting to retrieve sets of defining elements for the style through co-analysis~\cite{xu2010style,hu2017co,yu2018semi} or focusing on curve styles of an object silhouette~\cite{li2013curve,lun2015elements}. 
However, only a limited number of works proposed techniques to perform style transfer from one shape to another~\cite{lun2016functionality,ma2014analogy,cao2020psnet}. 
In the pre-deep learning era, proposed techniques either relied on tabu search to iteratively update a sample to make it perceptually similar to the chosen style reference while preserving its functionality,~\cite{lun2016functionality} or on recasting style transfer in terms of shape analogies~\cite{ma2014analogy} and correspondence transfer~\cite{han20153d}.
%
%
More recently, Yin \etal~\cite{yin2019logan} propose a generative approach to 3D shape transform. 
Each shape is encoded to a unique code and the transformation from one 3D object to another is left to a translator network.
The learned shape transformations are thus deterministic and style and content spaces are not treated separately.
Cao~\etal~\cite{cao2020psnet} propose \psnet{}, a point cloud equivalent of the original image style transfer algorithm~\cite{gatys2016image}.
\psnet{} extrapolates content and style representations as the intermediate activations and the corresponding Gram matrices of a pre-trained \pointnet{}~\cite{simonyan2014very} network. 
%
%
The output shape is generated by iteratively updating a random input until its content and style representations match the reference ones. 
Since network weights are frozen and optimization only affects the initial random input, their method cannot learn general translation functions and optimization must run for each different input pair.
Consequently, their method is much slower than a single forward pass and their explicit style encoding is not controllable, often resulting in a representation limited to encoding a spatial extent instead of embedding finer details.
%
%
%
\section{Method} \label{sec:method}
\begin{figure*}[t]
\begin{center}
\resizebox{0.9\linewidth}{!}{%
    \includegraphics[width=\textwidth,scale=0.5,trim={0.0cm 4.4cm 0.0cm 0.0cm},clip]{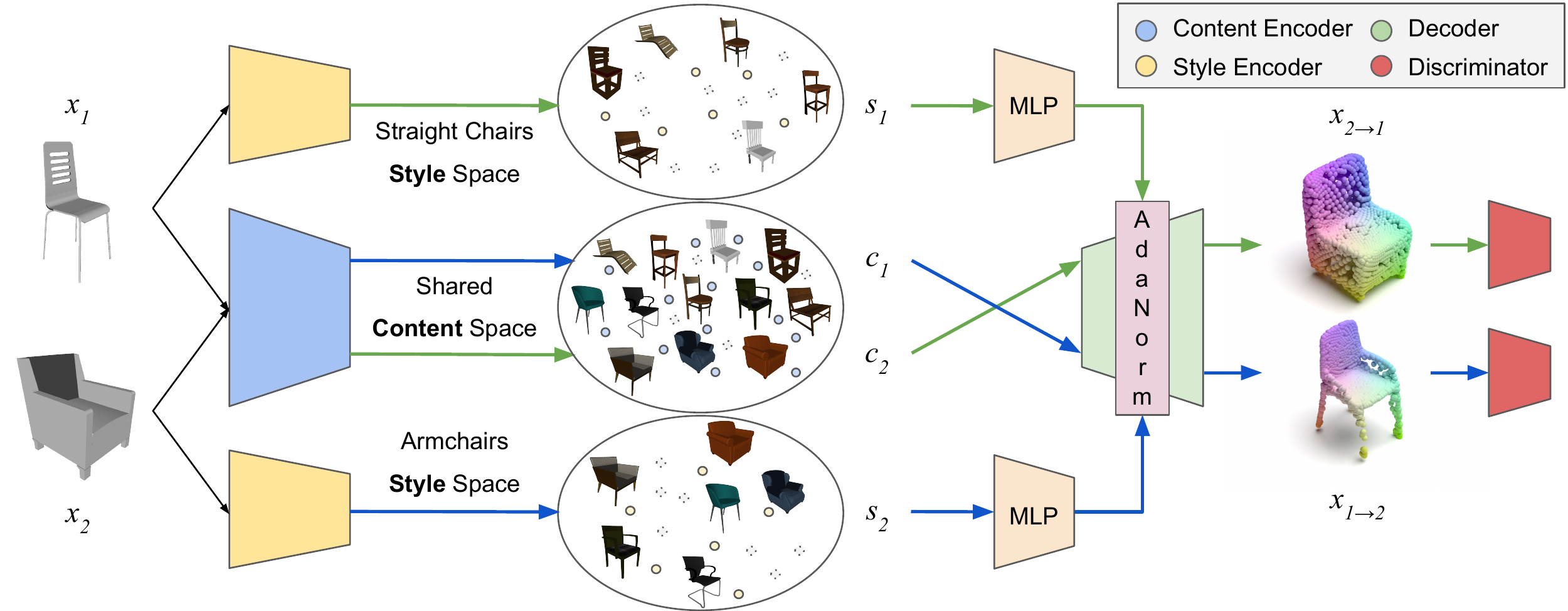} 
}
\end{center}
\vspace{-2mm}
    \caption{The \methodName{} architecture leverages a shared content encoder and two domain-specific style encoders to map the input families into content and style latent spaces respectively. A decoder is used to reconstruct 3D objects from the chosen content and style codes. Style codes are processed by an MLP to infer the center and scale parameters of each adaptive normalization layer in the decoder. To generate realistic outputs, we also adversarially train a discriminator for each style family.
    We visualize here a specific use case (\chairsPair{}) for the sake of explanation, though the approach can be applied to generic classes (see Section \ref{sec:experiments}). 
    }
    \vspace{-3mm}
    \label{img:method}
\end{figure*}

%
We now introduce \methodName{}, our framework for 3D style transfer.
%
%
In \Cref{ssec:assumptions}, we enumerate the assumptions under which it operates.
We then describe in \Cref{ssec:architecture} the proposed architecture.
In \Cref{ssec:3dsnet}, we depict the training strategy and loss components.
In \Cref{ssec:3dsnet_multi}, we extend our method to learn multimodal style distributions.

\subsection{Notation and Definitions} \label{ssec:assumptions}
Let $\domainOne{}$ and $\domainTwo{}$ be two domains of 3D shapes (\eg \textit{straight chair} and \textit{armchair}).
Similarly to the image-to-image translation problem, shape-to-shape translation can be tackled with or without supervision.
In the \textit{supervised} setting, paired samples $(\shapeOne{}, \shapeTwo{})$ from the joint distribution $\pOneTwo{}$ are given.
In the \textit{unsupervised} one, the given shapes $\shapeOne{}$ and $\shapeTwo{}$ are unpaired and independently sampled from the distributions $\pOne{}$ and $\pTwo{}$. 
Our proposal addresses the more general unsupervised setting, thus easily reducible to the paired one.
%
%
As shown in~\Cref{img:method}, we make a \textit{partially shared latent space} assumption, assuming that each data space $\mathcal{X}_i$ of domain $i$ can be projected into a shared content latent space $\contentSpace{}$ and an independent style latent space $\styleSpace{}_i$.
This assumption already proved effective for images~\cite{huang2018multimodal}.
A shape $x_i$ from the family $\mathcal{X}_i$, \ie $x_i \sim P_{\mathcal{X}_i}$, can be represented by a content code $c_i \in \contentSpace_i (=\contentSpace)$ shared across domains and a domain-specific style code $s_i \in \styleSpace{}_i$.

Our goal of transferring the style from one domain to another is equivalent to estimating the conditional distributions $\pOneCondTwo{}$ and $\pTwoCondOne{}$ via learned shape-to-shape translation models, $\pLearnedOneCondTwo{}$ and $\pLearnedTwoCondOne{}$ respectively. 
Generative adversarial autoencoders~\cite{makhzani2015adversarial} have been successfully used to learn data distributions.
We propose to learn for each family $\mathcal{X}_i$ encoding functions $\contentEncoder_i=\contentEncoder$ and $\styleEncoder_i$ to the shared content space $\contentSpace$ and domain-specific style space $\styleSpace_i$, respectively.
A corresponding decoding function $\globalDecoder{}_i$ can be used to recover the original shapes from their latent coordinates $(c_i, s_i)$ as $x_i=\globalDecoder{}_i(c_i, s_i)=\globalDecoder{}_i(\contentEncoder(x_i), \styleEncoder_i(x_i))$.
We can thus achieve domain translation by combining the style of one sample with the content representation of another through style-specific decoding functions, \eg $\shapeOneTwo{}=\globalDecoder{}_2(\contentEncoder(\shapeOne{}), \styleEncoder_2(\shapeTwo{}))$.
%
%
For this to work, we require that cycle-consistency~\cite{zhu2017unpaired,kim2017learning} holds, meaning that the input shape can be reconstructed by translating back the translated input shape.
This has proven effective to enforce that encoding and decoding functions across all domains are consistent with each other.
%

%
%
\subsection{Architecture} \label{ssec:architecture}
The proposed architecture can operate both on point clouds and meshes.
As illustrated in \Cref{img:method}, it takes as input a pair of independently sampled shapes $\shapeOne{}$, $\shapeTwo{}$, and solves both tasks of 3D reconstruction and style transfer.
The model, which must satisfy all the aforementioned assumptions in \Cref{ssec:assumptions}, comprises a handful of components: 
%
%
a shared content encoder $\contentEncoder{}$, a pair of domain-specific style encoders $\styleEncoder_1$ and $\styleEncoder_2$, a content decoder with weights shared across domains but made style-dependent thanks to adaptive normalization layers~\cite{huang2017arbitrary,li2018adaptive}, and a pair $(M_1, M_2)$ of domain-specific mapping functions \mbox{$M_i: \styleSpace_i \longrightarrow{} \mathrm{(\Gamma \times B)}$} from style coordinates $s_i$ to the pair of adaptive normalization scale and center parameters $(\gamma, \beta)$. 
Moreover, we rely on two domain-specific discriminators $\discriminator{}_i$ to adversarially train our generative model.

The overall architecture is thus an autoencoder through which we estimate both reconstruction and style translation functions.
In the following we describe and motivate the design choices for each component of our architecture.
Further details are provided in the supplementary material.
%

\noindent \textbf{Content and Style Encoders.}
To map the input shapes $\shapeOne{}$ and $\shapeTwo{}$ to the low-dimensional content and style latent spaces $\contentSpace{}$, $\styleSpace{}_1$, and $\styleSpace{}_2$, we leverage a shared content encoder $\contentEncoder{}$ and two domain-specific style encoders $\styleEncoder_1$ and $\styleEncoder_2$.
We adopt the \pointnet{}~\cite{qi2017pointnet} architecture for both the content and style encoders.
%
\pointnet{} has proven effective to learn point cloud embeddings in a permutation invariant way thanks to the use of max pooling operations. 
%
%

\noindent \textbf{Decoder.}
After encoding a shape $x_i$ to the shared content space $\contentSpace{}$ and to the style space $\styleSpace{}_i$ of its domain $i$, we obtain a low-dimensional representation of the input, composed of the disentangled content and style coordinates $(c_i, s_i)$.
Current state-of-the-art decoder architectures~\cite{groueix2018papier,yang2018foldingnet,yang2019pointflow,gupta2020neural} for meshes or point clouds usually rely on a single shape embedding $z \in \mathcal{Z}$ from a unique latent space $\mathcal{Z}$ as input.
To handle our disentangled representation $(c_i,s_i)$, we propose to replace standard normalization layers with Adaptive Normalization (\adanorm{})~\cite{huang2017arbitrary,li2018adaptive} layers to make these decoders style-dependent.
\adanorm{} layers allow to learn the center $\beta$ and scale $\gamma$ parameters of a normalization layer, \eg batch~\cite{ioffe2015batch} or instance normalization~\cite{ulyanov2016instance}, while conditioning them on the style code. 
Depending on the given style code, the Gaussian distribution to which the layer's output is normalized changes to mimic that of the corresponding training domain.
Different domains can indeed be described as different distributions in the deep activation space~\cite{segu2020batch}.
This comes with the advantage of conditioning the decoder on both content and style representations, thus enabling style transfer.

We validate our approach with a popular decoder for meshes and point clouds, \atlasnet~\cite{groueix2018papier}, and a recent but promising decoder, \meshflow~\cite{gupta2020neural}.
%
%
Since the content code $c_i$ embeds a shared and general representation of the input shape, we treat the content as the main input feature to feed such decoders, replacing the unique shape latent code $z$. 
Intermediate activations of the network are then conditioned on the style code $s_i$ via \adanorm{} layers.
Specifically, to learn the center $\beta$ and scale $\gamma$ parameters of each \adanorm{} layer, we feed the style code $s_i$ to a mapping function $M_i$. 
Moreover, since the style codes $s_1$ and $s_2$ come from the two different distributions $P_{\mathcal{S}_1}$ and $P_{\mathcal{S}_2}$, we must learn two mapping functions from each style space to a same space of \adanorm{} parameters, \ie $M_i: \styleSpace{}_i \longrightarrow{} \mathrm{(\Gamma \times B)}$, $\forall i \in \{1,2\}$.
We propose to implement each mapping function with a MLP made of two blocks, each composed of a fully connected layer followed by a dropout layer~\cite{srivastava2014dropout} and ReLU activation~\cite{nair2010rectified}. 
%
While standard \atlasnet{} and \meshflow{} architectures learn the mapping $P_{\mathcal{X}|\mathcal{Z}}$ from the unique latent space $\mathcal{Z}$ to the data space $\mathcal{X}$, our style-adaptive design choice 
%
%
allows to learn the data distribution of a domain $i$ given the corresponding content and style spaces $\contentSpace{}$ and $\styleSpace{}_i$, \ie $P_{\mathcal{X}_i|\contentSpace{},\styleSpace{}_i}$.
Specifically, each of these distributions is learned via a global decoding function $\globalDecoder{}_i = \globalDecoder{}(x_i|c_i,s_i)$ implemented as $\globalDecoder{}_i(c_i, s_i) = \contentDecoder{}(c_i, \mappingFunction{}_i(s_i))$, where $\contentDecoder{}$ is the chosen content decoder architecture conditioned on the style code by \adanorm{} layers.
Furthermore, to leave the basic functionality of \atlasnet{} and \meshflow{} decoders unchanged, their normalization layers are replaced by the corresponding adaptive counterpart, \ie \adabn{} and \adain{} respectively.
This proposal enables a single decoder to learn the reconstruction and translation functions for both input domains, respectively $\phi_{1\xrightarrow{}1}=\globalDecoder{}_1(c_1, s_1)=\globalDecoder{}_1(\contentEncoder(x_1), \styleEncoder_1(x_1))$ and $\phi_{1\xrightarrow{}2}=\globalDecoder{}_2(c_1, s_2)=\globalDecoder{}_2(\contentEncoder{}(x_1), \styleEncoder_2(x_2))$ for the family $\mathcal{X}_1$. Similar models can be deducted for $\mathcal{X}_2$.

\noindent \textbf{Discriminator.}
To adversarially train our generative model we exploit a discriminator $\discriminator{}_i$ for each domain $i$ and implement them as \pointnet{} feature extractors with final MLP.

\subsection{\methodName{}: 3D Style Transfer} \label{ssec:3dsnet}
We now describe the loss functions and the training procedure employed in our framework.
First, we adopt a \textit{reconstruction loss} to ensure that the encoders and decoders are each other's inverses.
Next, an \textit{adversarial loss} matches the  distribution of the translated shapes with the distribution of shapes in the target domain.
Finally, a \textit{cycle-reconstruction loss} enforces that encoding and decoding functions for all domains are consistent with each other.
%

\noindent\textbf{3D Reconstruction Loss.}
Given a latent representation $(c_i, s_i)$ of a 3D shape $x_i$ from the family $\mathcal{X}_i$, one of the decoder's goals is to reconstruct the surface of the shape.
Let $\mathcal{T}$ be a set of points sampled on the surface of the target mesh $x_i$, and $\mathcal{Y}_{i,i}$ the set of 3D vertices generated by our model $\phi_{i\xrightarrow{}i}(x_i)$.
We then minimize the \textit{bidirectional Chamfer distance} between the two point sets:
\begin{equation*}
    \mathcal{L}^{x_{i}}_{\text{rec}} = \mathbb{E}_{\mathcal{T},\mathcal{Y}_{i,i}} [\sum_{p\in \mathcal{T}} \min_{q \in \mathcal{Y}_{i,i}} || p - q ||^2 +  \sum_{q\in \mathcal{Y}_{i,i}} \min_{p \in \mathcal{T}} || p - q ||^2 ] 
    \label{eq:reconstruction_loss}
\end{equation*}

\noindent\textbf{Adversarial Loss.}
We exploit generative adversarial networks (GANs) to match the distribution of the translated shapes with the distribution of shapes in the target domain.
Given a discriminator $\discriminator{}_i$ that aims at distinguishing between real shapes and translated shapes in the domain $i$, our model should be able to fool it by generating shapes that are indistinguishable from real ones in the target domain.
%
%
To this end, we introduce the adversarial loss $\mathcal{L}^{x_{j}}_{\text{adv}}$:
\begin{align*}
	\mathcal{L}^{x_{j}}_{\text{adv}} = &\mathbb{E}_{c_{i}, s_{j}}[\log(1-\discriminator{}_{j}(H_{j}(c_{i},s_{j})))] + \mathbb{E}_{x_{j}}[\log \discriminator{}_{j}(x_{j})] 
\end{align*}
where $i$ and $j$ are two different training domains.

\noindent\textbf{3D Cycle-Reconstruction Loss.}
For our method to work, cycle-consistency must hold, meaning that we can, for example, reconstruct the input shape $x_1$ by translating back the translated input shape, \ie $x_{1\xrightarrow{}2\xrightarrow{}1}=\phi_{2\xrightarrow{}1}(\phi_{1\xrightarrow{}2}(x_1))$.
Let $\mathcal{U}$ be a set of points sampled on the surface of $x_i$, and $\mathcal{Y}_{i, j, i}$ the set of 3D vertices of the surface $x_{i\xrightarrow{}j\xrightarrow{}i}$ generated by our cycle model $\phi_{j\xrightarrow{}i}(\phi_{i\xrightarrow{}j}(x_i))$.
Similarly to $\mathcal{L}_{\text{rec}}$, we define $\mathcal{L}_{\text{cycle}}$ as the Chamfer loss between these two point sets.

\noindent\textbf{Total Loss.}
The final objective is a weighted sum of all aforementioned components.
We jointly train the encoders $\contentEncoder{}$, $\styleEncoder_1$, $\styleEncoder_2$, the decoder $\contentDecoder{}$, the mapping functions $\mappingFunction{}_1$, $\mappingFunction{}_2$ and the discriminators $\discriminator{}_1$ and $\discriminator{}_2$ to optimize the total loss:
\begin{align*} 
\underset{\substack{\contentEncoder{}, \styleEncoder_1, \styleEncoder_2, \\ \contentDecoder{}, \mappingFunction{}_1, \mappingFunction{}_2}}\min\ \underset{\discriminator{}_1, \discriminator{}_2}\max\ \mathcal{L}_{\text{tot}} = \lambda_{\text{rec}}(\mathcal{L}^{x_{1}}_{\text{rec}} + \mathcal{L}^{x_{2}}_{\text{rec}})&\ + \notag \\
	\lambda_{\text{adv}}(\mathcal{L}^{x_{1}}_{\text{adv}}+\mathcal{L}^{x_{2}}_{\text{adv}}) + 
	\lambda_{\text{cycle}}(\mathcal{L}^{x_{1}}_{\text{cycle}}+\mathcal{L}^{x_{2}}_{\text{cycle}})&
\end{align*} \label{eq:total_loss}
where $\lambda_{\text{rec}}$, $\lambda_{\text{adv}}$, $\lambda_{\text{cycle}}$ are the weights for each loss term.

\subsection{\methodNameMultimodal{}: Learning a Multimodal Style Space} \label{ssec:3dsnet_multi}
Multimodal image-to-image translation~\cite{huang2018multimodal,patashnik2020balagan,lee2018diverse,nizan2020breaking} was recently proposed to learn non-deterministic image translation functions.  
Since the reconstruction and translation functions learned through \methodName{} are deterministic, we introduce a variant that allows to learn the multimodal distribution over the style space, enabling for the first time non-deterministic multimodal shape-to-shape translation.
We name the proposed extension \methodNameMultimodal{}, which empowers \methodName{} by enabling it to solve both (i) \textit{example-driven style transfer} and (ii) \textit{multimodal shape translation to a given style family}.
%

If we assume a prior distribution over the style spaces and learn to encode and decode it, we can sample points from each of the two assumed distributions $\hat{P}_{\styleSpace{}_1}$ and $\hat{P}_{\styleSpace{}_2}$ to generate a much larger variety of generated samples for a given input content reference.
This would also allow to learn a non-deterministic translation function, for which, depending on the points sampled from the style distribution, we can generate different shapes.
To achieve this goal, in addition to the previously presented losses, we take inspiration from the MUNIT~\cite{huang2018multimodal} proposal for images and rely on additional \textit{latent consistency losses} to learn the underlying style distributions.
%
%
During inference, the learned multimodal style distribution allows us to generate multiple shape translations from a single sample $x$.
In particular, we can extrapolate the content $c$ from $x$, sample random points $s_1 \sim \hat{P}_{\styleSpace{}_1} = \mathcal{N}(0,\identity)$ or $s_2 \sim \hat{P}_{\styleSpace{}_2} = \mathcal{N}(0,\identity)$, and decode them through the global decoding function $\globalDecoder{}_i$ of the preferred style family $i$. 

\noindent\textbf{Latent Reconstruction Loss.}
Since style codes are sampled from a prior distribution, we need a latent reconstruction loss to enforce that decoders and encoders are inverse with respect to the style spaces:
\begin{align*}
	\mathcal{L}^{c_{1}}_{\text{rec}} = \mathbb{E}_{c_{1}\sim p(c_{1}), s_{2}\sim q(s_{2})}[||\contentEncoder{}(\globalDecoder{}_{2}(c_{1},s_{2}))-c_{1}||_{1}]\\
	\mathcal{L}^{s_{2}}_{\text{rec}} = \mathbb{E}_{c_{1}\sim p(c_{1}), s_{2}\sim q(s_{2})}[||\styleEncoder_2(\globalDecoder{}_{2}(c_{1},s_{2}))-s_{2}||_{1}]
\end{align*}
where $q(s_{2}) \sim \mathcal{N}(0, \identity)$ is the style prior, \mbox{$p(c_{1})$ is given by $c_{1} = \contentEncoder{}(x_{1})$ and $x_{1}\sim \pOne{}$.}
The corresponding $\mathcal{L}^{c_{2}}_{\text{rec}}$ and $\mathcal{L}^{s_{1}}_{\text{rec}}$ losses are defined similarly.

\noindent\textbf{Total Loss.}
The final objective is a weighted sum of the aforementioned components, including those in \Cref{ssec:3dsnet}:
\begin{align*} 
\underset{\substack{\contentEncoder{}, \styleEncoder_1, \styleEncoder_2, \\ \contentDecoder{}, \mappingFunction{}_1, \mappingFunction{}_2}}\min\ \underset{\discriminator{}_1, \discriminator{}_2}\max\ \mathcal{L}_{\text{tot}} = \lambda_{\text{rec}}(\mathcal{L}^{x_{1}}_{\text{rec}} + \mathcal{L}^{x_{2}}_{\text{rec}})&\ + \notag \\
	\lambda_{\text{adv}}(\mathcal{L}^{x_{1}}_{\text{adv}}+\mathcal{L}^{x_{2}}_{\text{adv}}) + 
	\lambda_{\text{cycle}}(\mathcal{L}^{x_{1}}_{\text{cycle}}+\mathcal{L}^{x_{2}}_{\text{cycle}})&\ + \notag \\
	\lambda_{c}(\mathcal{L}^{c_{1}}_{\text{rec}}+\mathcal{L}^{c_{2}}_{\text{rec}})+\lambda_{s}(\mathcal{L}^{s_{1}}_{\text{rec}}+\mathcal{L}^{s_{2}}_{\text{rec}})&
\end{align*} \label{eq:total_loss_multi}
where $\lambda_{\text{rec}}$, $\lambda_{\text{adv}}$, $\lambda_{\text{cycle}}$, $\lambda_{\text{c}}$, $\lambda_{\text{s}}$ are the weights for each loss.
%
\section{Experiments} \label{sec:experiments}
We validate the proposed framework and show how the content space encodes the shared underlying structure, while each style latent space nicely embeds the distinctive traits of its domain.
For non-rigid models (\eg animals), we also show how this disentangled representation allows the content latent space to implicitly learn a pose parametrization shared between two style families.

\subsection{Evaluation Metrics}
\noindent \textbf{3D Perceptual Loss Network.}
The perceptual similarity, known as \lpips{}~\cite{johnson2016perceptual}, is often computed as a distance between two images in the VGG~\cite{simonyan2014very} feature space.
To compute a perceptual distance $d_{i,j}$ between two point clouds $x_i$ and $x_j$ given a pre-trained network $F$, we introduce the \threedlpips{} metric.
Inspired by the 2D-\lpips{} metric, we leverage as feature extractor $F$ a \pointnet{}~\cite{qi2017pointnet} pre-trained on the \shapenet{}~\cite{chang2015shapenet} dataset.
We compute deep embeddings $(\mathbf{f}_i, \mathbf{f}_j)=(F^l(x_i),F^l(x_j))$ for the inputs $(x_i,x_j)$, the vector of the intermediate activations at each layer $l$. We then normalize the activations along the channel dimension, and take the $L_2$ distance between the two resulting embeddings. 
%
%
We then average across the spatial dimension and across all layers to obtain the final \threedlpips{} metrics.

\noindent \textbf{Style Transfer Score.}
We introduce a novel metric to evaluate style transfer approaches in general, and we name it \textit{Style Transfer Score} (\metricName{}).
Given two input shapes $(\shapeOne, \shapeTwo)$ from two different domains, we want to evaluate the effectiveness of the transfer of style from $\shapeTwo$ to $\shapeOne$. 
Each shape can be represented as a point in the \pointnet{} feature space.
We deem style transfer as successful when the augmented sample $x_{1\xrightarrow{}2}$ is perceptually more similar to the target $\shapeTwo$ than to the source $\shapeOne$.
%
%
%
Let $\Delta_{\text{source}} = |d_{\text{source},\text{aug}}| - |d_{\text{source},\text{rec}}|$ and $\Delta_{\text{target}} = |d_{\text{aug},\text{target}}| - |d_{\text{rec},\text{target}}|$, where $d_{i,j}$ is the previously introduced \threedlpips{} metrics.
If, from a perceptual similarity point of view, the augmented sample $x_{1\xrightarrow{}2}$ is farther from the source $\shapeOne{}$ than the source reconstruction $x_{1\rightarrow1}$, \ie $(|d_{\text{source},\text{aug}}| > |d_{\text{source},\text{rec}}| \Rightarrow \Delta_{\text{source}} > 0)$, while also closer to the target $\shapeTwo{}$ than the target reconstruction $x_{2\rightarrow2}$  $(|d_{\text{aug},\text{target}}| < |d_{\text{rec},\text{target}}| \Rightarrow \Delta_{\text{target}} < 0)$, then style transfer can be considered successful.
Following this principle, we introduce \metricName{} to validate style transfer effectiveness by weighting $\Delta_{\text{source}}$ positively and $\Delta_{\text{target}}$ negatively.
\begin{align*}
    \metricName{}=\Delta_{\text{source}}-\Delta_{\text{target}}
\end{align*}
For an intuitive visualization of the proposed metric, we refer the reader to the supplementary material.
%
\begin{figure}[t]

\begin{center}
\resizebox{0.9\columnwidth}{!}{%
    \includegraphics[width=\linewidth,scale=0.5,trim={0.0cm 0.5cm 5.8cm 0.0cm},clip]{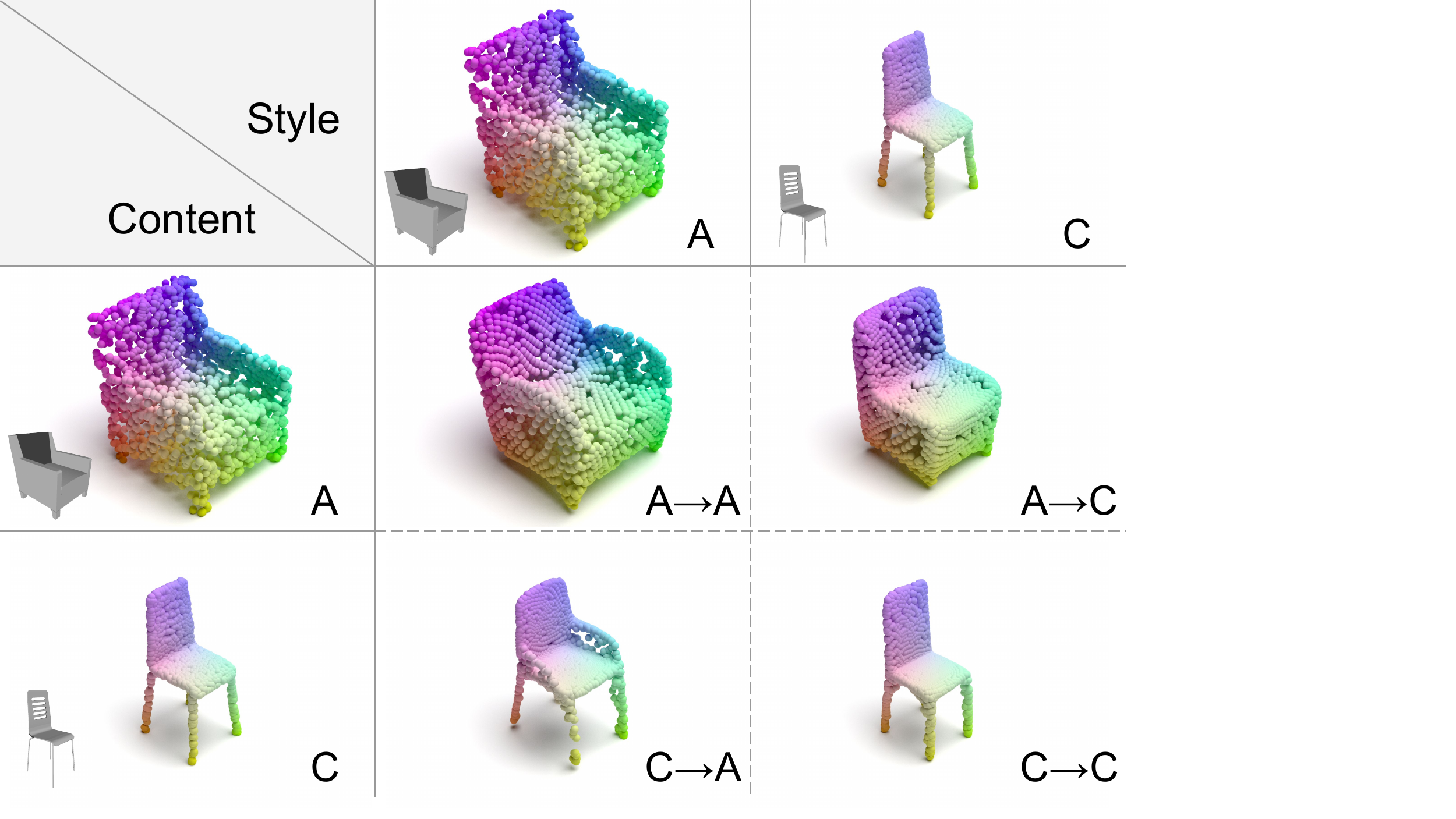} 
}
\end{center}
    \caption{\methodName{} reconstruction (main diagonal) and style transfer (anti-diagonal) results on \textit{armchair} (A) $\leftrightarrow$ \textit{straight chair} (C).}
    \label{img:chairs_style}
    \vspace{-3mm}
\end{figure}

%
\begin{figure*}[t]
\vspace{-3mm}
\begin{center}
    \includegraphics[width=0.85\textwidth,trim={0.0cm 6.8cm 0.0cm 0.0cm},clip]{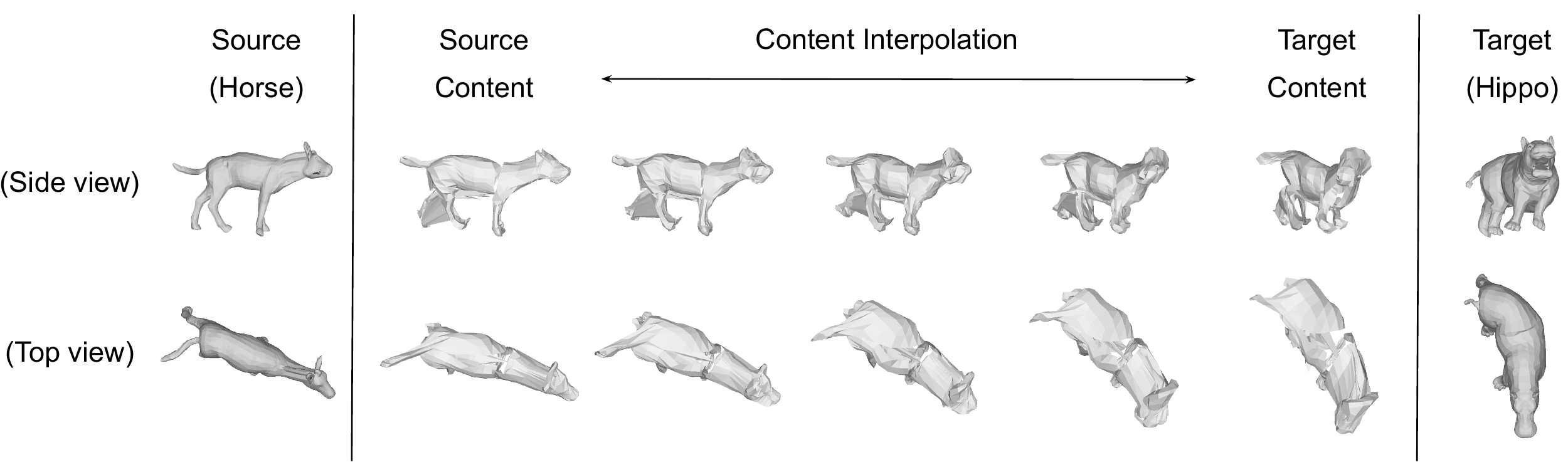} 
\end{center}
\vspace{-3mm}
    \caption{Latent walk in the content space learned by \methodName{} (Ours) for the \smalPair{} dataset. We generate shapes by uniformly interpolating between source and target contents, while keeping source style \horses{} constant. Decoder is \atlasnet{} with \adanorm{} layers.}
    \vspace{-3mm}
    \label{img:content_animals}
\end{figure*}

%
\noindent \textbf{Average \threedlpips{} Distance.}
To evaluate the variability of the generated shapes, we follow~\cite{huang2018multimodal} and randomly select $100$ samples for each style family. For each sample, we generate $19$ output pairs by combining the same content code with two different style codes randomly sampled from the same style distribution of a given family.
We then compute the \threedlpips{} distance between the samples in each pair, and average over the total $1\,900$ pairs. 
The larger it is, the more output diversity the adopted method achieves.

\subsection{Baselines}
\noindent \textbf{AdaNorm.}
We refer to our framework trained only with the reconstruction loss, and without the adversarial and cycle consistency losses, as \adanorm{}.

\noindent \textbf{3DSNet + Noise.}
As baseline for the multimodal setting, we compare against \methodName{} with random noise\footnote{We apply a Gaussian noise with standard deviation $\sigma$ on the style codes. We empirically select $\sigma$ to be the maximum possible value allowing realistic results (\ie $0.1$ with \atlasnet, $1.0$ with \meshflow).}.

\subsection{Experimental Setting}
One of the main limitations in the development of 3D style transfer approaches is the lack of ad-hoc datasets.
We propose to leverage pairs of subcategories of established benchmarks for 3D reconstruction, \ie \shapenet{}~\cite{chang2015shapenet} and the \smal{}~\cite{zuffi20173d} datasets.
\shapenet{} is a richly-annotated, large-scale dataset of 3D shapes, containing 51\,300 unique 3D models and covering 55 common object categories.
The \smal{} dataset is created by fitting the SMPL~\cite{loper2015smpl} body model to different animal categories.

For the chosen pairs to satisfy a definition of style for 3D shapes~\cite{xu2010style,dosovitskiy2015learning} and the partially shared latent space assumption in \Cref{ssec:assumptions}, they must share a common underlying semantic structure.
We validate hence our approach on two of the most populated object categories in \shapenet{} and identify for each two subcategories as corresponding stylistic variations: \chairs{} (\chairsPair{}), \planes{} (\planesPair{}).
We further test our method on the categories \hippos{} and \horses{} in \smal{}.
\setlength{\tabcolsep}{1.5pt}
\begin{table}[t]
\begin{center}
\resizebox{1.04\columnwidth}{!}{%
    \begin{tabular}{ c | c | c c c | c c | c }
    \toprule[1.5pt]
    Categories & Method & $\mathcal{L}_{\textbf{rec}}$ & $\mathcal{L}_{\textbf{adv}}$ & $\mathcal{L}_{\textbf{cycle}}$ & $\Delta_{s}$ & $\Delta_{t}$ & \metricName \\
    \midrule[1.5pt]
    \multirow{3}{*}{\chairsPair{}} & \adanorm{}                     & \cmark & \xmark & \xmark & 3.91 & -8.23 & 12.14 \\
    \cmidrule{2-8}
                                   & \multirow{2}{*}{\methodName{}} & \cmark & \cmark & \xmark & 2.61 & \textbf{-11.11} & 13.72 \\
                                   &                                & \cmark & \cmark & \cmark & \textbf{3.45} & -10.96 & \textbf{14.41} \\
    \midrule[1.5pt]
    \multirow{3}{*}{\planesPair{}} & \adanorm{}                     & \cmark & \xmark & \xmark & 2.32 & -11.86 & 14.18 \\
    \cmidrule{2-8}
                                   & \multirow{2}{*}{\methodName{}} & \cmark & \cmark & \xmark & 3.35 & -8.66 & 12.00 \\
                                   &                                & \cmark & \cmark & \cmark & \textbf{3.37} & \textbf{-12.60} & \textbf{15.97} \\
    \midrule[1.5pt]
    \multirow{3}{*}{\smalPair{}}   & \adanorm{}                     & \cmark & \xmark & \xmark & -0.06 & -0.68 & 0.62 \\
    \cmidrule{2-8}
                                   & \multirow{2}{*}{\methodName{}} & \cmark & \cmark & \xmark & 1.24 & -1.23 & 2.47 \\
                                   &                                & \cmark & \cmark & \cmark & \textbf{2.11} & \textbf{-1.63} & \textbf{3.74} \\
    \bottomrule[1.5pt]
    \noalign{\smallskip}
    \end{tabular}
} 
\caption{Style Transfer Score (\metricName{}) for different variants of \methodName{} on the \chairsPair{}, \planesPair{}, and \smalPair{} datasets. \atlasnet{} with \adanorm{} layers is used as decoder. Results are multiplied by $10^2$.}
\label{table:sts_all}
\vspace{-7mm}
\end{center}
\end{table}

\subsection{Evaluation of \methodName{}}
We evaluate the effectiveness of our style transfer approach on the proposed benchmarks by comparing the proposed \metricName{} metric across different variants of our approach.

\noindent \textbf{\shapenet{}.}
We validate our approach on two different families of \chairs{} from the \shapenet{} dataset, \ie \textit{armchairs} and \textit{straight chairs}, having $1\,995$ and $1\,974$ samples respectively.
For the \planes{} category, we choose the domains \textit{jet} and \textit{fighter aircraft}, with $1\,235$ and $745$ samples.

\Cref{img:chairs_style} shows qualitative results of the reconstruction and translation capabilities of our framework on the pair \chairsPair{}. 
Our proposal excels in transforming \textit{armchairs} in \textit{simple chairs} and vice versa, while accurately maintaining the underlying structure. 
\methodName{} successfully confers armrests to straight chairs and removes them from armchairs, producing novel realistic 3D shapes.
%
%

\begin{table}[]
\small
\begin{center}
    \begin{tabular}{ c c | c c | c }
    \toprule
    Method & Decoder & $\Delta_{s}$ & $\Delta_{t}$ & \metricName \\
    \midrule
    \multirow{2}{*}{\adanorm{}} & \atlasnet & 3.91 & -8.23 & 12.14 \\
                             & \meshflow & 3.55 & -4.86 & 8.41 \\
    \midrule
    \multirow{2}{*}{\methodName{}} & \atlasnet & 3.45 & -10.96 & 14.41 \\
                            & \meshflow & \textbf{6.19} & \textbf{-20.66} & \textbf{26.85} \\
    \bottomrule
    \noalign{\smallskip}
    \end{tabular}
\caption{Ablation study on the decoder architecture. We compare the Style Transfer Score (\metricName{}) for different variants of our method on the \chairsPair{}. Results are multiplied by $10^2$.}
\label{table:chairs_meshflow}
\end{center}
\vspace{-7mm}
\end{table}

\begin{figure*}[t]
\vspace{-3mm}
\begin{center}
    \includegraphics[width=0.84\textwidth,trim={0.0cm 6.6cm 0.0cm 0.0cm},clip]{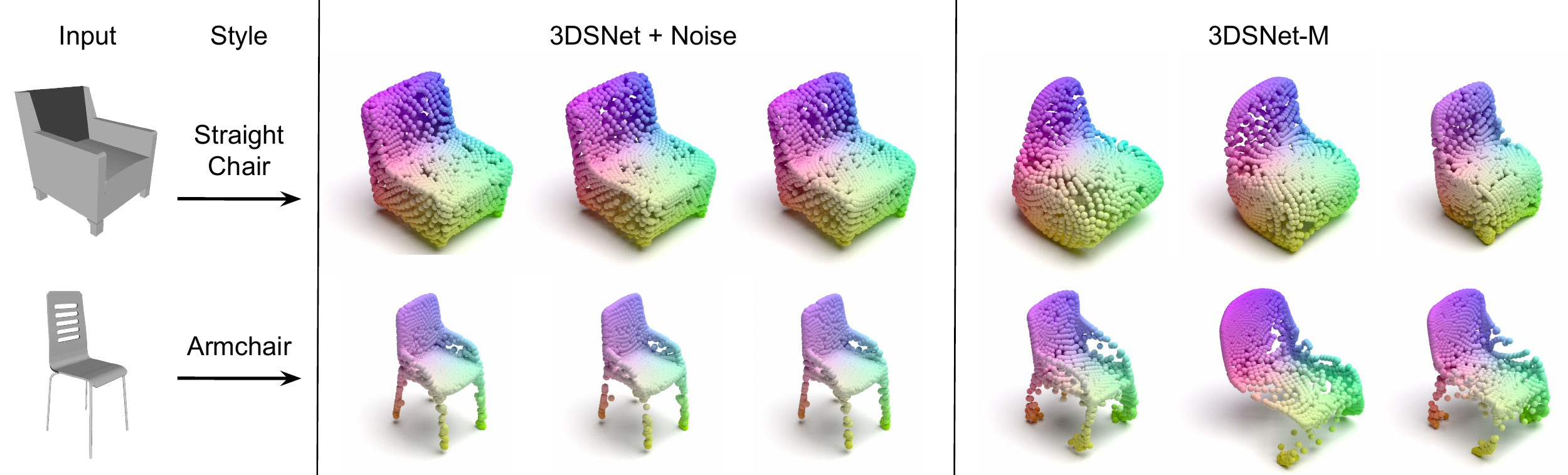} 
\end{center}
\vspace{-3mm}
    \caption{Comparison of multimodal shape translation for \methodName{} + Noise and \methodNameMultimodal{} on \chairsPair{}. \methodNameMultimodal{} produces a larger diversity of samples, while \methodName{} learns to disregard the noise added on the style code.}
    \vspace{-4mm}
    \label{img:multimodal_chairs}
\end{figure*}


\Cref{table:sts_all} illustrates quantitative evaluation on the \chairs{} and \planes{} categories, where \methodName{} consistently outperforms the \adanorm{} baseline.
Moreover, results validate the assumption that cycle-consistency is required to constrain all model encoders and decoders to be inverses, giving a further boost on the proposed metric.
For the \chairs{} category, we also include an ablation study (\Cref{table:chairs_meshflow}) comparing the \metricName{} of both \adanorm{} and \methodName{} with two different decoders: \atlasnet{} and \meshflow{}, both with adaptive normalization layers.
%
%
Quantitative results highlight the superiority of \methodName{} with \meshflow{} decoder on the \chairs{} benchmark.
This can be explained by the different kind of normalization layer in the two decoders: while \atlasnet{} relies on batch normalization, \meshflow{} leverages instance normalization, which has already proven more effective for style transfer tasks~\cite{ulyanov2016instance} in the image domain.

\noindent \textbf{Non-rigid Shapes: \smal{}.}
We test \methodName{} also on the \hippos{} and \horses{} categories from \smal{}, a dataset of non-rigid shapes.
%
%
For each category, we generate $2\,000$ training samples and $200$ validation samples, sampling the model parameters from a Gaussian distribution with variance $0.2$.

%
\Cref{table:sts_all} presents quantitative evaluation via the \metricName{} for different variants of \methodName{}.
We then qualitatively validate the goodness of the learned latent spaces by taking a latent walk in the learned content space, interpolating across source and target content representations. 
\Cref{img:content_animals} shows uniformely distanced interpolations of the content codes of the two input samples.
It can be observed that the learned content space nicely embeds an implicit pose parametrization, and the shapes produced with the interpolated contents display a smooth transition between the pose of the source sample and that of the target, while maintaining the style of the source family.
This is expected, since content features are forced to be shared across the different domains, respecting the shared pose parametrization between the domains \hippos{} and \horses{}.
\subsection{Evaluation of \methodNameMultimodal{}}
Following \Cref{ssec:3dsnet_multi}, we evaluate the variability of the samples generated with \methodNameMultimodal{} compared to the baseline \methodName{} + Noise.
%
%
\Cref{table:chairs_variance} shows the Average \threedlpips{} distance for the \chairsPair{} dataset, where results show how \methodNameMultimodal{} outperforms the baseline by a margin of more than $100\%$.
Moreover, our method with adaptive \meshflow{} as decoder performs extremely well on \chairs{}, allowing a further improvement of an order of magnitude over the \atlasnet{} counterpart.

\setlength{\tabcolsep}{1.5pt}
\begin{table}[h]
\begin{center}
\resizebox{0.7\columnwidth}{!}{%
    \begin{tabular}{ c c c }
    \toprule
    Method & Decoder & Avg. \threedlpips{} \\
    \midrule
    \methodName{} + Noise & \multirow{2}{*}{\atlasnet{}} & 0.347 \\
    \methodNameMultimodal{} &                              & 0.630 \\
    \midrule
    \methodName{} + Noise & \multirow{2}{*}{\meshflow} & 6.010 \\
    \methodNameMultimodal{} &                             & \textbf{14.251} \\
    \bottomrule
    \noalign{\smallskip}
    \end{tabular}
}
\caption{Average \threedlpips{} for different variants of our method on \chairsPair{}. Results are multiplied by $10^3$.}
\label{table:chairs_variance}
\end{center}
\vspace{-5mm}
\end{table}

\Cref{img:multimodal_chairs} highlights how the proposed \methodNameMultimodal{} generates a larger variety of translated shapes compared to the baseline \methodName{} + Noise, for which the network learns to disregard variations on the style embedding.

\subsection{Comparison with \psnet{}}
\begin{figure}[t]
\begin{center}
    \includegraphics[width=0.9\linewidth,scale=0.5,trim={0.0cm 4.8cm 9.0cm 0.0cm},clip]{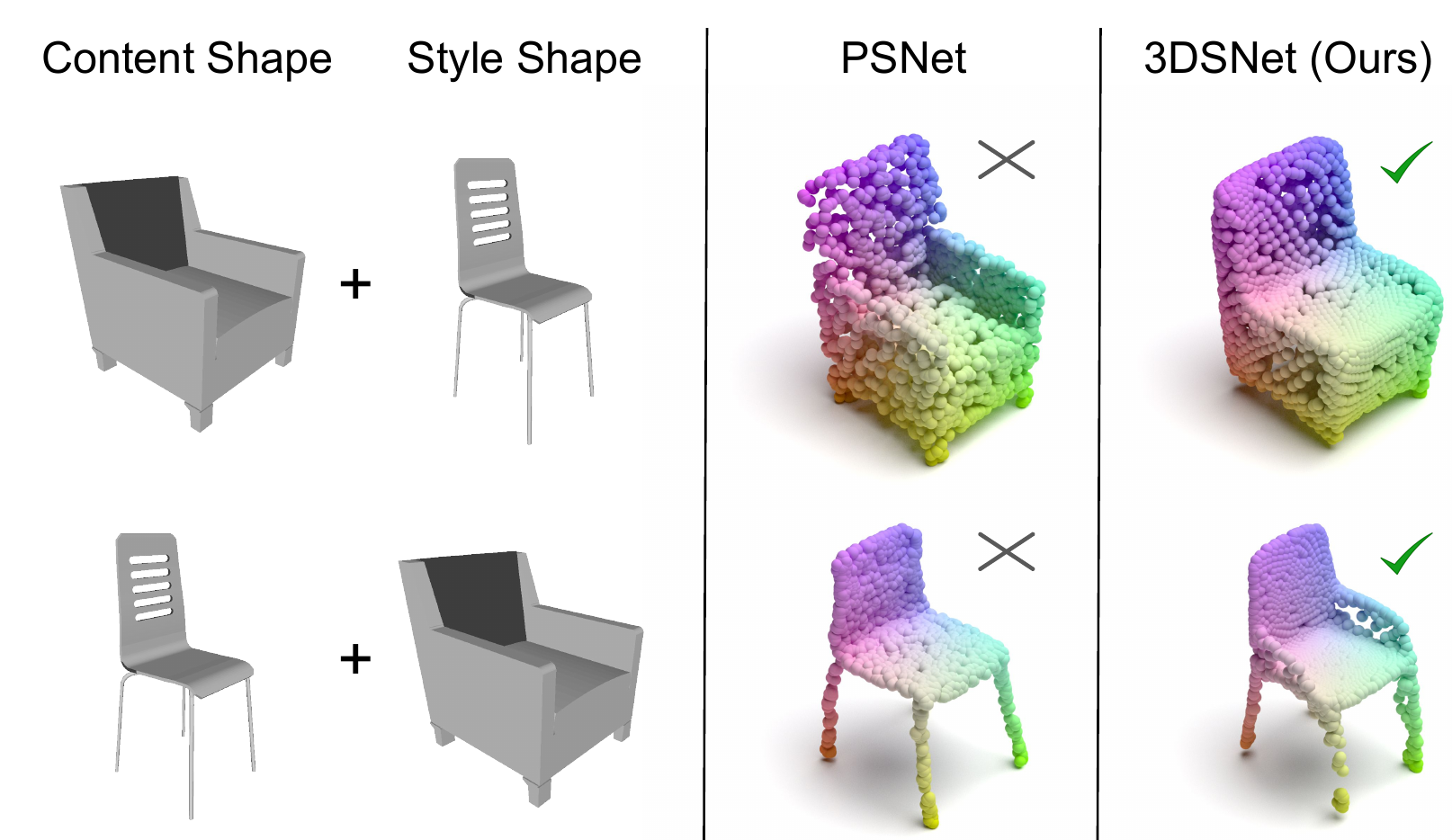} 
\end{center}
    \vspace{-3mm}
    \caption{Qualitative comparison of style transfer effectiveness on \chairsPair{} for \methodName{} (Ours) and its main competitor \psnet{}~\cite{cao2020psnet}.}
    \vspace{-5mm}
    \label{img:psnet_chairs}
\end{figure}

%
While \psnet{}~\cite{cao2020psnet} tackles the style transfer problem in the 3D setting, it is a gradient-based technique that operates on point clouds. 
\Cref{img:psnet_chairs} illustrates the advantages of the proposed \methodName{} over \psnet{}'s explicit style representation, which appears to limit understanding of style to the spatial extent of a point cloud, thus failing to translate a shape from one domain to another, \ie Armchair into Straight Chair, and vice versa.
In contrast,  \methodName{} effortlessly extends to both meshes and point clouds, while achieving style transfer and content preservation, \ie armrests added or removed while maintaining the underlying structure.
%
\section{Conclusions} \label{sec:conclusions}
We propose \methodName{}, the first generative 3D style transfer approach to learn disentangled content and style spaces.
Our framework simultaneously solves the shape reconstruction and translation tasks, even allowing to learn implicit parametrizations for non-rigid shapes.
The multimodal extension enables translating into a diverse set of outputs.

Nevertheless, our proposal inherits limitations from 3D reconstruction approaches.
The loss of details noticeable in reconstructed shapes is arguably due to the loss of high frequency components caused by the max pooling layer in \pointnet{}~\cite{wang2020softpoolnet}.
Moreover, optimizing the Chamfer distance between point sets promotes meshes with nicely fitting vertices but stretched edges, as in~\Cref{img:content_animals}.
%
%
Such limitations emphasize the need for improvements in 3D reconstruction.
%

In its novelty, our work opens up several unexplored research directions, such as 3D style transfer from a single view or style-guided deformations of existing templates.

\clearpage
{\small
\bibliographystyle{ieee_fullname}
\bibliography{egbib}
}
\clearpage

\newcommand{\refColor}{blue}
\newcommand{\Sref}[1]{\textcolor{\refColor}{#1}}

\section{Supplementary Material}
We provide supplementary material to validate our method.
\Cref{ssec:sts} provides further details on the proposed Style Transer Score;
\textit{N.B.:} Blue references point to the original manuscript.

\subsection{Style Transfer Score} \label{ssec:sts}
\begin{figure}[t]
\begin{center}
    \includegraphics[width=\linewidth,scale=0.4,trim={0.0cm 1.2cm 8.5cm 0.0cm},clip]{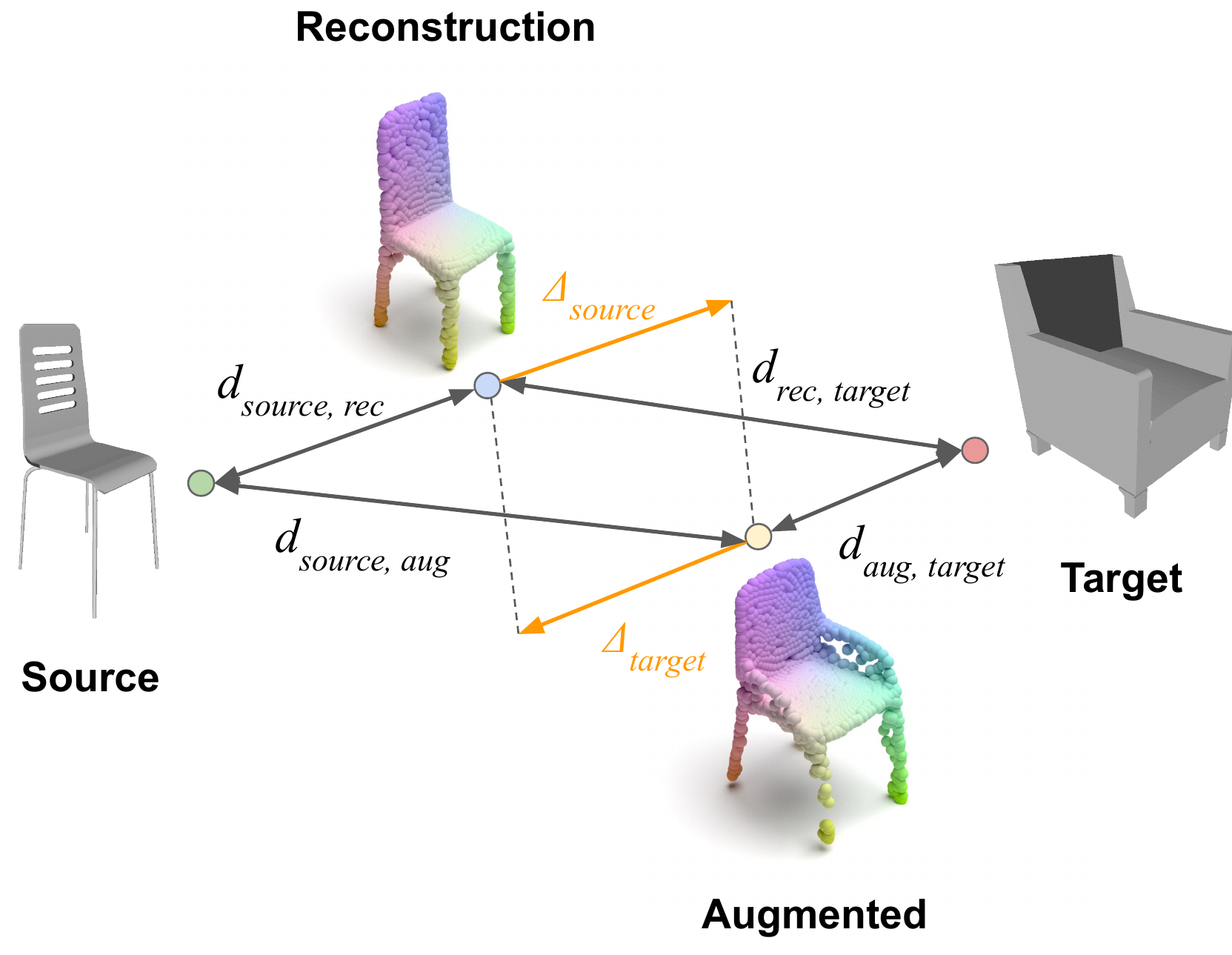} 
\end{center}
    \caption{Illustration in the \pointnet{} feature space of source and target samples, reconstruction and augmented shapes for the task \chairsPair{}. $\metricName{}=\Delta_{source}-\Delta_{target}$.}
    \label{img:sts}
\end{figure}

We here expand the definition of the Style Transfer Score (\metricName{}) proposed in \Sref{Section 4.1}.
Since perceptual similarity (\threedlpips{}) is evaluated as a distance between deep features in the \pointnet{} feature space, we can identify any input sample as a point in the \pointnet{} feature space.
In \Cref{img:sts}, we can see how, by mapping in the feature space the source samples, the target samples, the source reconstructions and augmented samples, we can compute the \metricName{} following the procedure depicted in \Sref{Section 4.1}:
\begin{align*}
    \metricName{}=\Delta_{\text{source}}-\Delta_{\text{target}}.
\end{align*}
The metric inherently takes into account the quality of the 3D reconstruction, providing a useful hint to recognize effective style transfer methods among those that simultaneously perform reconstruction.
It is worth noting that the proposed metric can be extended to any setting, \eg images, text, 3D processing, by simply changing the adopted feature extractor.

\subsection{Implementation Details}
To train \methodName{}, \methodNameMultimodal{} and all their variants, we select the same set of parameters for all possible tasks, configurations and with both \atlasnet{}~\cite{groueix2018papier} and \meshflow{}~\cite{gupta2020neural} decoders.

Our framework is trained for a total 180 epochs, with an initial learning rate $10^{-3}$ that is decayed by a factor $10^{-1}$ at epochs $120$, $140$ and $145$.
We adopt Adam~\cite{kingma2014adam} as optimizer and train with a batch size of $16$.
We choose as dimension of the content code $1024$ and of the style code $512$.
The bottleneck size for the \pointnet{} discriminators is also $1024$, followed by an MLP with $3$ blocks of fully-connected, dropout and ReLU layers.

As adversarial loss, we choose Least Squares Generative Adversarial Networks (GAN), which proved to address instability problems that are frequent in GANs.
Concerning the weights attributed to each loss components, \methodName{} is trained with $\lambda_{\text{rec}}=1$, $\lambda_{\text{adv}}=0.1$, $\lambda_{\text{cycle}}=1$.
The multimodal variant \methodNameMultimodal{} adds the latent reconstruction components with weights $\lambda_{\text{c}}=\lambda_{\text{s}}=0.1$.

\subsection{Evaluation of 3D Reconstruction}
%

%
\begin{table}[t]
\begin{center}
    \begin{tabular}{ c | c c c |  c }
    \toprule
    Method & $\mathcal{L}_{\textbf{rec}}$ & $\mathcal{L}_{\textbf{adv}}$ & $\mathcal{L}_{\textbf{cycle}}$ & $CD$ \\
    \midrule
    \adanorm{} & \cmark & \xmark & \xmark & \textbf{7.38} \\
    \midrule
    \multirow{2}{*}{\methodName{}} & \cmark & \cmark & \xmark & 7.57 \\
                            & \cmark & \cmark & \cmark & 7.63 \\
    \bottomrule
    \noalign{\smallskip}
    \end{tabular}
\caption{Chamfer distance (CD) comparison for different variants of \methodName{} on the \chairsPair{} dataset. \atlasnet{} with \adanorm{} layers is used as decoder. Results are multiplied by $10^3$. \adanorm{} is the baseline leveraging only the reconstruction loss. With \methodName{}, Chamfer distance does not increases significantly, meaning that we manage to build on reconstruction techniques while not negatively affecting their performance.}
\label{table:chamfer}
\end{center}
\vspace{-5mm}
\end{table}

Our framework relies on state-of-the-art 3D reconstruction techniques.
We here evaluate quantitatively how our style transfer approach affects the reconstruction performance.
Ideally, adding style transfer capabilities to a reconstruction approach should not affect the reconstruction performance.

In \Cref{table:chamfer}, the evaluation of the Chamfer distance for different variants of our framework proves how the multiple added losses do not affect significantly the quality of traditional reconstruction approaches, which simply train with a reconstruction loss only, nominally the bidirectional Chamfer distance.


%
\subsection{Extended Comparison with PSNet}
\begin{figure*}[t]
\vspace{-3mm}
\begin{center}
    \includegraphics[width=\textwidth,width=0.85\textwidth,trim={0.0cm 6.8cm 0.0cm 0.0cm},clip]{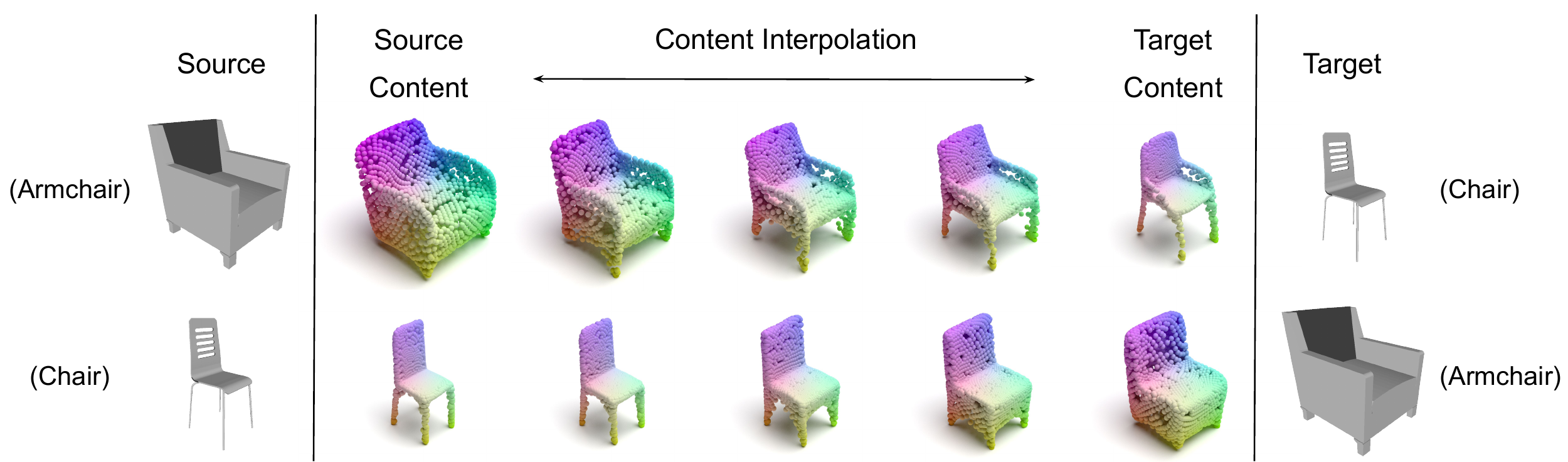} 
\end{center}
\vspace{-3mm}
    \caption{Latent walk in the content space learned by \methodName{} (Ours) for the \chairsPair{} dataset. We generate shapes by uniformly interpolating between source and target contents, while keeping the respective source style constant. Decoder is \meshflow{} with \adanorm{} layers.}
    \vspace{-3mm}
    \label{img:content_chairs}
\end{figure*}

%
\begin{table}[t]
\begin{center}
    \begin{tabular}{ c | c c | c }
    \toprule
    Method & $\Delta_{s}$ & $\Delta_{t}$ & \metricName \\
    \midrule
    PSNet~\cite{cao2020psnet} & 0.89 & -1.00 & 1.89  \\
    \midrule
    \adanorm{} \textit{(Ours)} & 3.91 & -8.23 & 12.14 \\
    \midrule
    \methodName{} \textit{(Ours)} & \textbf{3.45} & \textbf{-10.96} & \textbf{14.41}  \\
    \bottomrule
    \noalign{\smallskip}
    \end{tabular}
\caption{Comparison of Style Transfer Score (\metricName{}) for PSNet and \methodName{} on the \chairsPair{} dataset. \atlasnet{} with \adanorm{} layers is used as decoder. Results are multiplied by $10^2$.}
\label{table:chairs_sts_psnet}
\end{center}
\vspace{-5mm}
\end{table}

\setlength{\tabcolsep}{3pt}
\begin{table}[t]
\begin{center}
    \begin{tabular}{ c | c c | c }
    \toprule
    Method & $\Delta_{s}$ & $\Delta_{t}$ & \metricName \\
    \midrule
    PSNet~\cite{cao2020psnet} & 0.33 & -1.18 & 1.51  \\
    \midrule
    \adanorm{} \textit{(Ours)} & 2.32 & -11.86 & 14.18 \\
    \midrule
    \methodName{} \textit{(Ours)} & \textbf{3.37} & \textbf{-12.60} & \textbf{15.97}  \\
    \bottomrule
    \noalign{\smallskip}
    \end{tabular}
\caption{Comparison of Style Transfer Score (\metricName{}) for PSNet and \methodName{} on the \planesPair{} dataset. \atlasnet{} with \adanorm{} layers is used as decoder. Results are multiplied by $10^2$.}
\label{table:planes_sts_psnet}
\end{center}
\vspace{-5mm}
\end{table}

\begin{figure}[t]
\begin{center}
    \includegraphics[width=0.95\linewidth,scale=0.5,trim={0.0cm 5.3cm 9.0cm 0.0cm},clip]{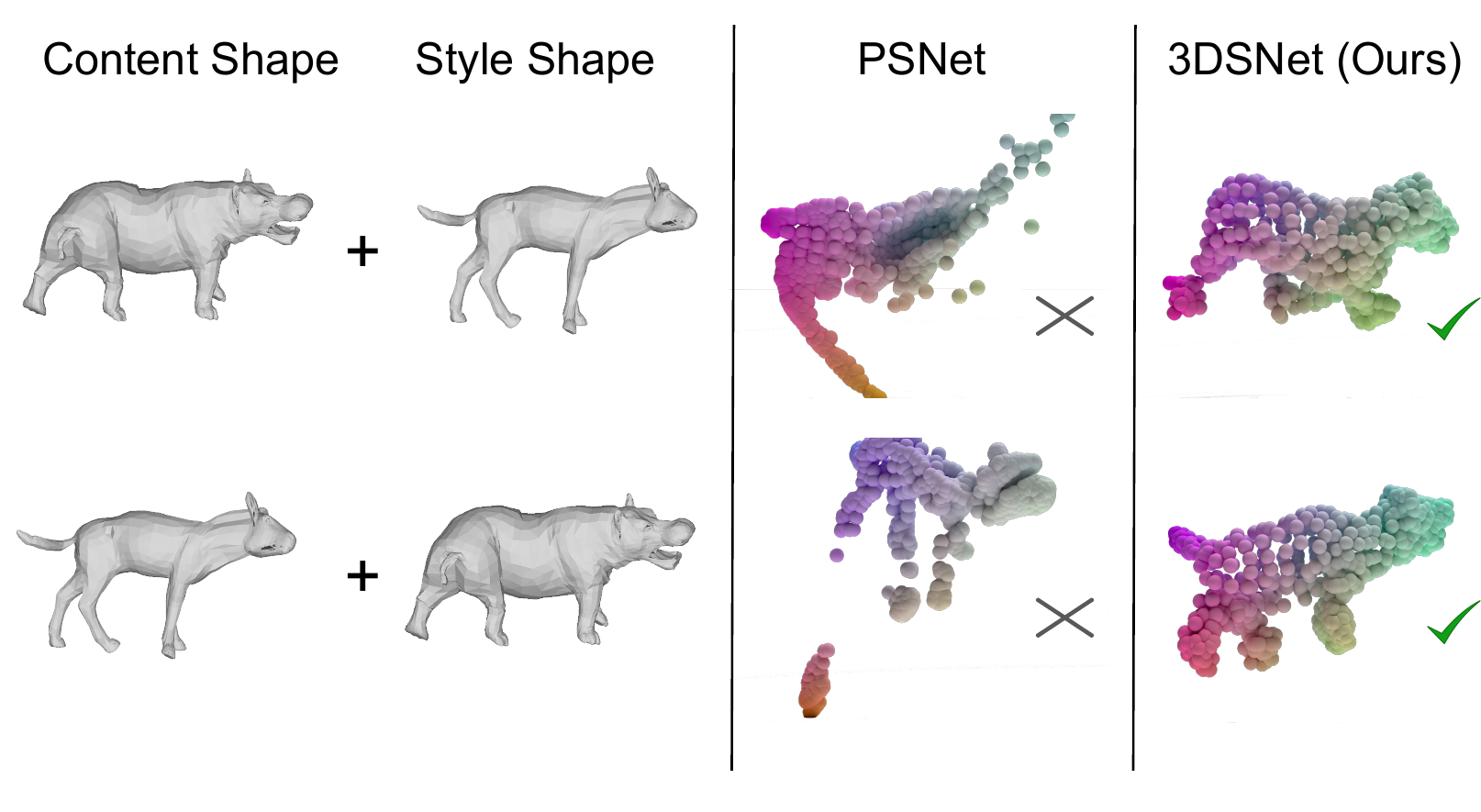} 
\end{center}
    \vspace{-5mm}
    \caption{Qualitative comparison of style transfer effectiveness on \smalPair{} for \methodName{} (Ours) and its main competitor \psnet{}~\cite{cao2020psnet}.}
    \vspace{-2mm}
    \label{img:psnet_smal}
\end{figure}

%
\begin{figure}[h]
\begin{center}
\resizebox{0.9\columnwidth}{!}{%
    \includegraphics[width=\linewidth,scale=0.5,trim={0.0cm 0.5cm 5.8cm 0.0cm},clip]{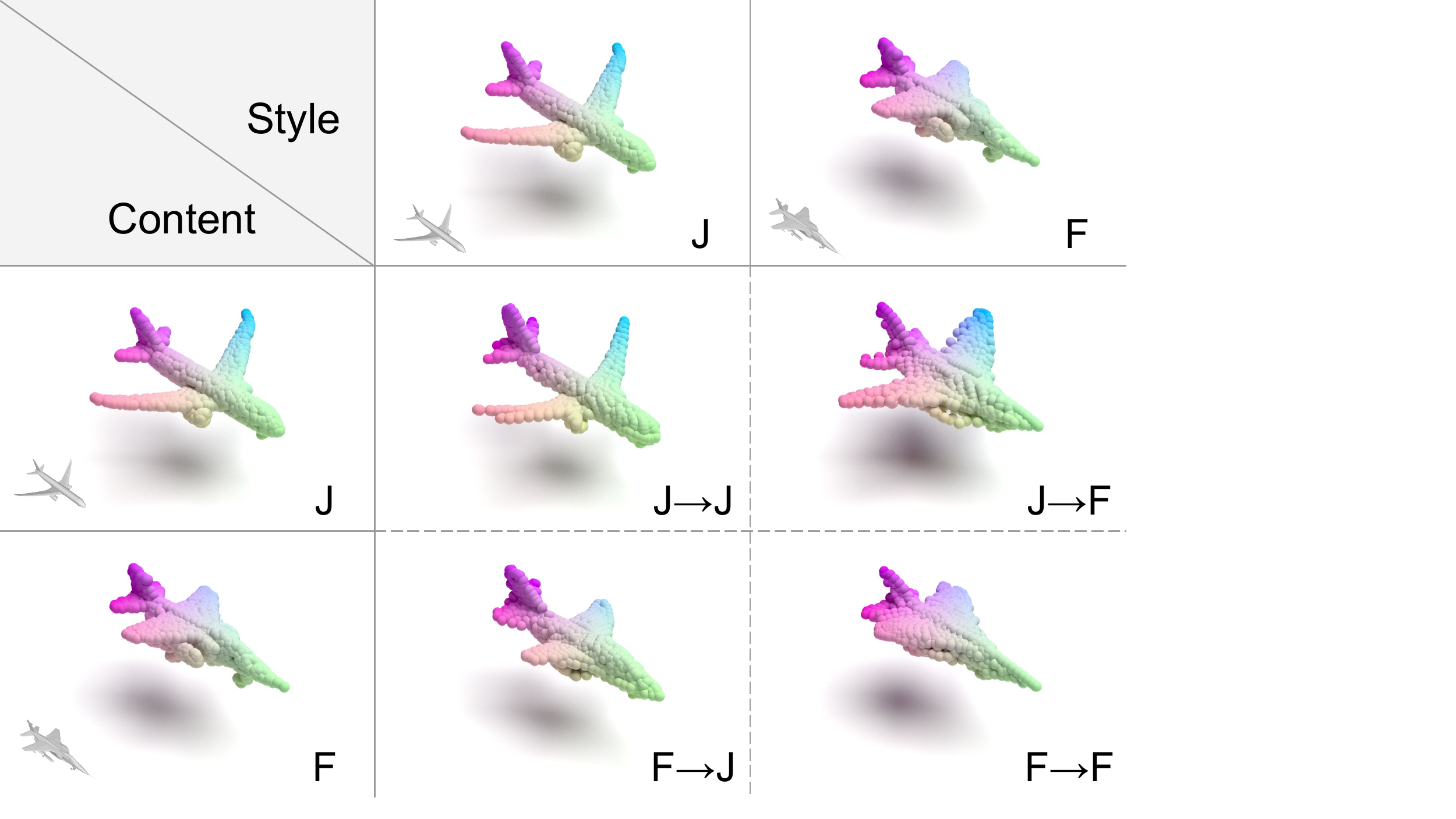} 
}
\end{center}
    \vspace{-3mm}
    \caption{\methodName{} reconstruction (main diagonal) and style transfer (anti-diagonal) results on \textit{jet} (J) $\leftrightarrow$ \textit{fighter aircraft} (F).}
    \label{img:planes_style}
    \vspace{-5mm}
\end{figure}

%
We extend the comparison with PSNet by validating the STS for PSNet against ours. 
\Cref{table:chairs_sts_psnet} and \Cref{table:planes_sts_psnet} prove the quality of our style transfer approach on the \chairsPair{} and \planesPair{} datasets, respectively.
\methodName{} outperforms PSNet on both benchmarks by over an order of magnitude.
This was expected, since \metricName{} evaluates the effectiveness of style transfer and relies on the \threedlpips{}, a perceptual similarity metric.
Indeed, PSNet already proved in \Sref{Figure 6} to limit its style transfer abilities to the spatial extent, hence not performing well from a perceptual perspective.
This lack of perceptual translation effectivess is reflected in the poor \metricName{} performance.

We also show a comparison of qualitative style transfer results obtained with both PSNet and \methodName{} on the \smal{} dataset.
\Sref{Figure 6} already shows how PSNet fails in encoding features that are peculiar of a given style category.
\Cref{img:psnet_smal} further shows how PSNet performs badly on the \smal{} dataset, failing to produce realistic results.

\subsection{Additional Examples}
\begin{figure*}[t]
\begin{center}
\resizebox{0.9\linewidth}{!}{%
    \includegraphics[width=\textwidth,scale=0.5,trim={0.0cm 17.5cm 0.0cm 0.0cm},clip]{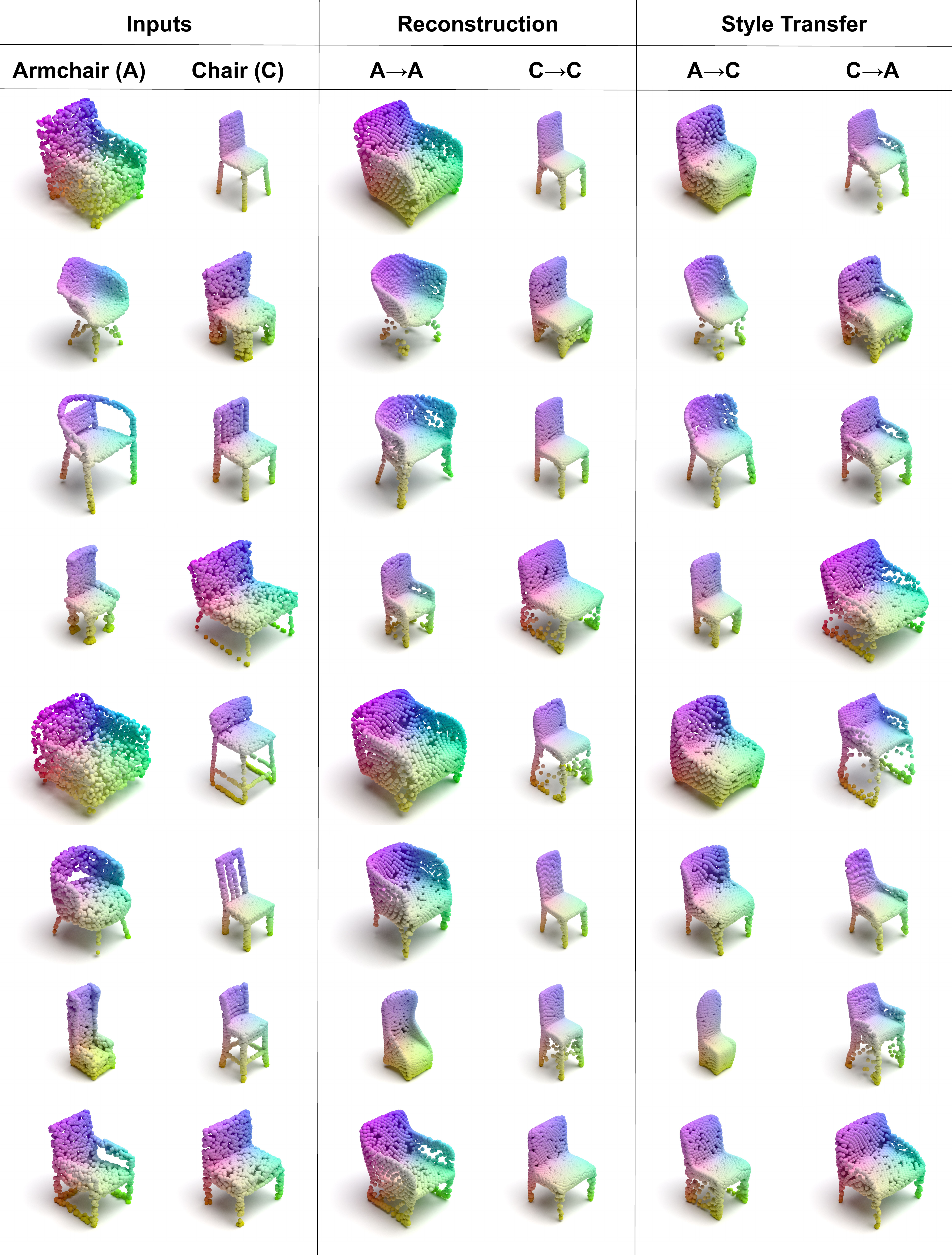} 
}
\end{center}
\vspace{-2mm}
    \caption{Visualization of reconstruction and style transfer results obtained by \methodName{} on $8$ pairs of the \chairsPair{} dataset. First two columns illustrate the inputs to our framework; second two columns show reconstructed shapes for each of the two inputs; last two columns depict style transfer results, where the notation $A\rightarrow{}B$ indicates that the shape from domain $A$ is translated to domain $B$ by combining the content of an example from $A$ with the style of an example from $B$. 
    }
    \vspace{-3mm}
    \label{img:chairs_supplementary}
\end{figure*}

\begin{figure*}[t]
\begin{center}
\resizebox{0.95\linewidth}{!}{%
    \includegraphics[width=\textwidth,scale=0.5,trim={0.0cm 22.0cm 0.0cm 0.0cm},clip]{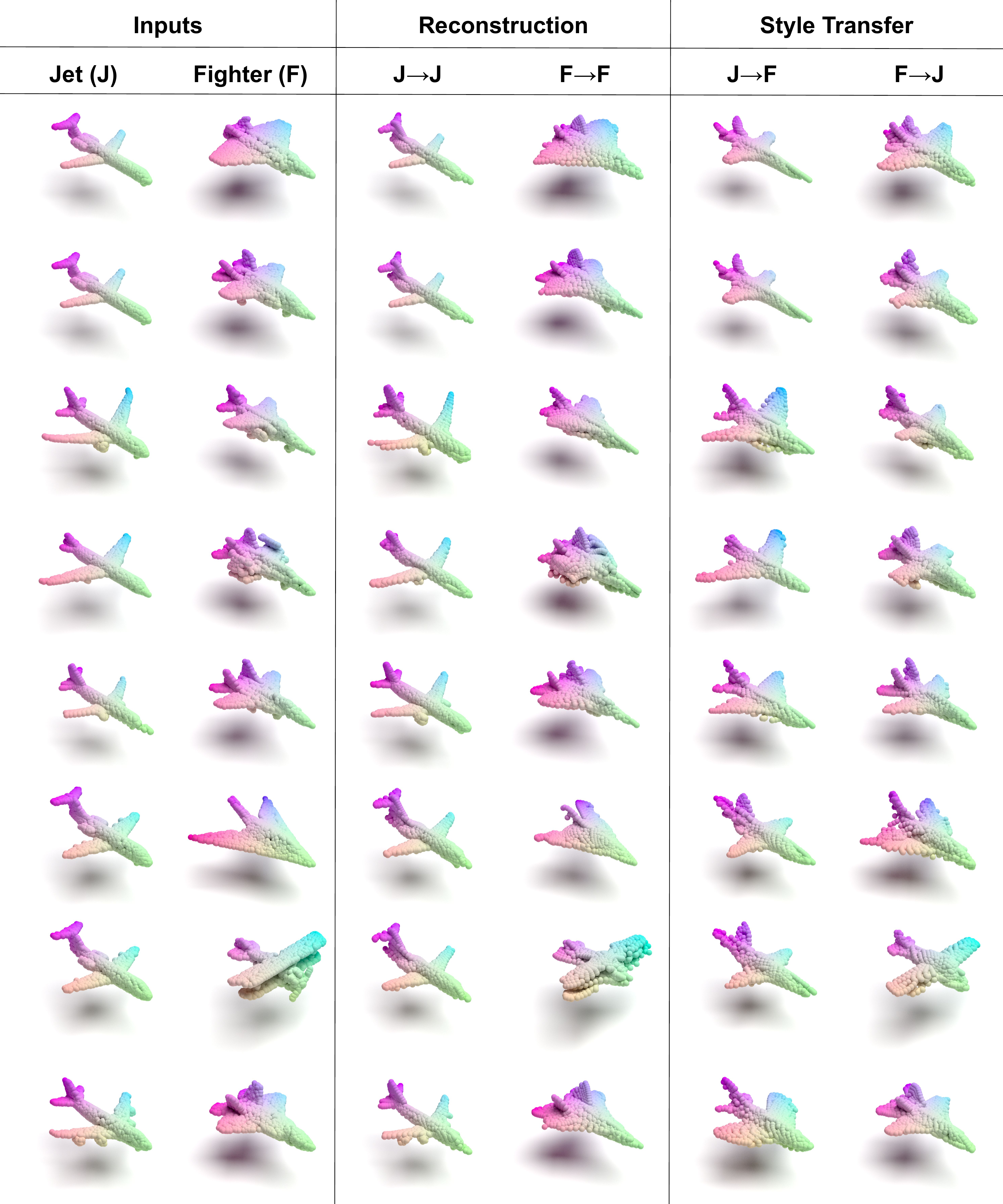} 
}
\end{center}
\vspace{-2mm}
    \caption{Visualization of reconstruction and style transfer results obtained by \methodName{} on $8$ pairs of the \planesPair{} dataset. First two columns illustrate the inputs to our framework; second two columns show reconstructed shapes for each of the two inputs; last two columns depict style transfer results, where the notation $A\rightarrow{}B$ indicates that the shape from domain $A$ is translated to domain $B$ by combining the content of an example from $A$ with the style of an example from $B$. 
    }
    \vspace{-3mm}
    \label{img:planes_supplementary}
\end{figure*}

%

\noindent\textbf{Chairs Latent Walk.}
Analogously to \Sref{Figure 4}, in \Cref{img:content_chairs} we show a latent walk in the content space learned by \methodName{} for the \chairsPair{} dataset.
Our method is able to keep the source style fixed (\ie with or without armrests), while accurately interpolating between the content of the source example and that of the reference target example.

\noindent\textbf{Chairs.}
In \Cref{img:chairs_supplementary}, we provide $8$ examples of the reconstruction and translation capabilities of \methodName{} on the \chairsPair{} dataset.

\noindent\textbf{Planes.}
In \Cref{img:planes_supplementary}, we provide $8$ examples of the reconstruction and translation capabilities of \methodName{} on the \planesPair{} dataset.
Furthermore, in \Cref{img:planes_style}, we remark how the model captures the stylistic differences in the dominant curves of \textit{jets} and \textit{fighter aircrafts}, successfully translating a \textit{jet} into a more aggressive \textit{fighter} and also managing the inverse transformation by smoothing the stark lines of the \textit{fighter aircraft}.

\end{document}